\documentclass{article}
\usepackage{spconf,amsmath,epsfig}
\usepackage{multirow}
\usepackage{dsfont}
\usepackage{stfloats}
\usepackage{subfig} 
\usepackage{tabu}  
\usepackage{cite}
\usepackage{booktabs}
\usepackage{algorithm}
\usepackage{algorithmic}
\usepackage{hyperref}

\title{Dual-Tasks Siamese Transformer Framework for Building Damage Assessment}
\name
 {Hongruixuan Chen$^{1,3}$, 
 Edoardo Nemni$^{1}$\sthanks{Corresponding Author (edoardo.nemni@unitar.org). This work was supported in part by the National Key R$\&$D Program
of China under Grant 2019YFE0126800. The work at the United Nations Satellite Centre (UNOSAT) is part of the operational mapping service funded by the Norwegian Ministry of Foreign Affairs. }, 
 Sofia Vallecorsa$^{2}$, 
 Xi Li$^{3}$,
 Chen Wu$^{3}$,
 Lars Bromley$^{1}$
}
 
 \address{$^1$ United Nations Satellite Centre (UNOSAT), United Nations Institute for \\ Training and Research (UNITAR) \\
 $^2$ European Organization for Nuclear Research (CERN)\\
 $^3$ State Key Laboratory of Information Engineering in Surveying, Mapping and Remote Sensing, \\
 Wuhan University
 }
 
\begin{document}
%
\maketitle
\begin{abstract}
Accurate and fine-grained information about the extent of damage to buildings is essential for humanitarian relief and disaster response. However, as the most commonly used architecture in remote sensing interpretation tasks, Convolutional Neural Networks (CNNs) have limited ability to model the non-local relationship between pixels. Recently, Transformer architecture first proposed for modeling long-range dependency in natural language processing has shown promising results in computer vision tasks. Considering the frontier advances of Transformer architecture in the computer vision field, in this paper, we present a Transformer-based damage assessment architecture (DamFormer). In DamFormer, a siamese Transformer encoder is first constructed to extract non-local and representative deep features from input multitemporal image-pairs. Then, a multitemporal fusion module is designed to fuse information for downstream tasks. Finally, a lightweight dual-tasks decoder aggregates multi-level features for final prediction. To the best of our knowledge, it is the first time that such a deep Transformer-based network is proposed for multitemporal remote sensing interpretation tasks. The experimental results on the  large-scale damage assessment dataset xBD demonstrate the potential of the Transformer-based architecture.
\end{abstract}
\section{Introduction}
\label{sec:intro}

\par Disasters such as hurricanes, earthquakes, floods, volcanic eruptions, and wildfires cause significant damage and economic losses every year. Timely and accurate building damage assessment is critical for humanitarian assistance, disaster response, and post-disaster reconstruction. Multitemporal very high-resolution satellite imageries are often used in the dual-task of identifying the extent of buildings and the severity of the damage. 

\par Deep learning, especially Convolutional Neural Networks (CNNs), has shown great results in computer vision and remote sensing applications including building damage assessment \cite{rs12162532,Chen2020TGRS,ZHENG2021112636}.  However, CNN lacks a global understanding of the images and the ability to model the non-local relationship between pixels. To capture long-range dependency for further performance improvement, some advanced modules, such as dilated convolution and self-attention mechanism were introduced \cite{chen2017deeplab,zhang2019self}. Nonetheless, in addition to a significant computational cost, these modules were often designed at the end of CNN-backbone, which implies that the relationship in low-level information was not captured.

\par Transformers architectures \cite{vaswani2017attention} have shown great performances in natural language processing (NLP) and their ability to model long-range dependencies had motivated researchers to explore its adaptation to computer vision tasks. In fact, the Vision Transformer (ViT) \cite{dosovitskiy2021an}  and its variants \cite{xie2021segformer,Liu_2021_ICCV} outperformed the state-of-the-art CNN in many computer vision tasks when trained on sufficient data. However, the Transformer-based architecture has not been investigated in large-scale multitemporal remote sensing interpretation tasks. More importantly, compared to computer vision tasks, multitemporal remote sensing interpretation tasks are subject to different complex challenges, such as larger object scale differences, and diverse image patterns caused by different sensor types and atmospheric conditions. 

\par In this paper, we explore the potential of Transformer-based backbones applying an end-to-end dual-tasks siamese Transformer architecture named DamFormer for the task of building damage assessment. We believe that this work can provide a different perspective to multitemporal remote sensing interpretation.

\section{Methodology}\label{sec:2}
\begin{figure*}[ht]

  \centering
  \includegraphics[scale=0.57]{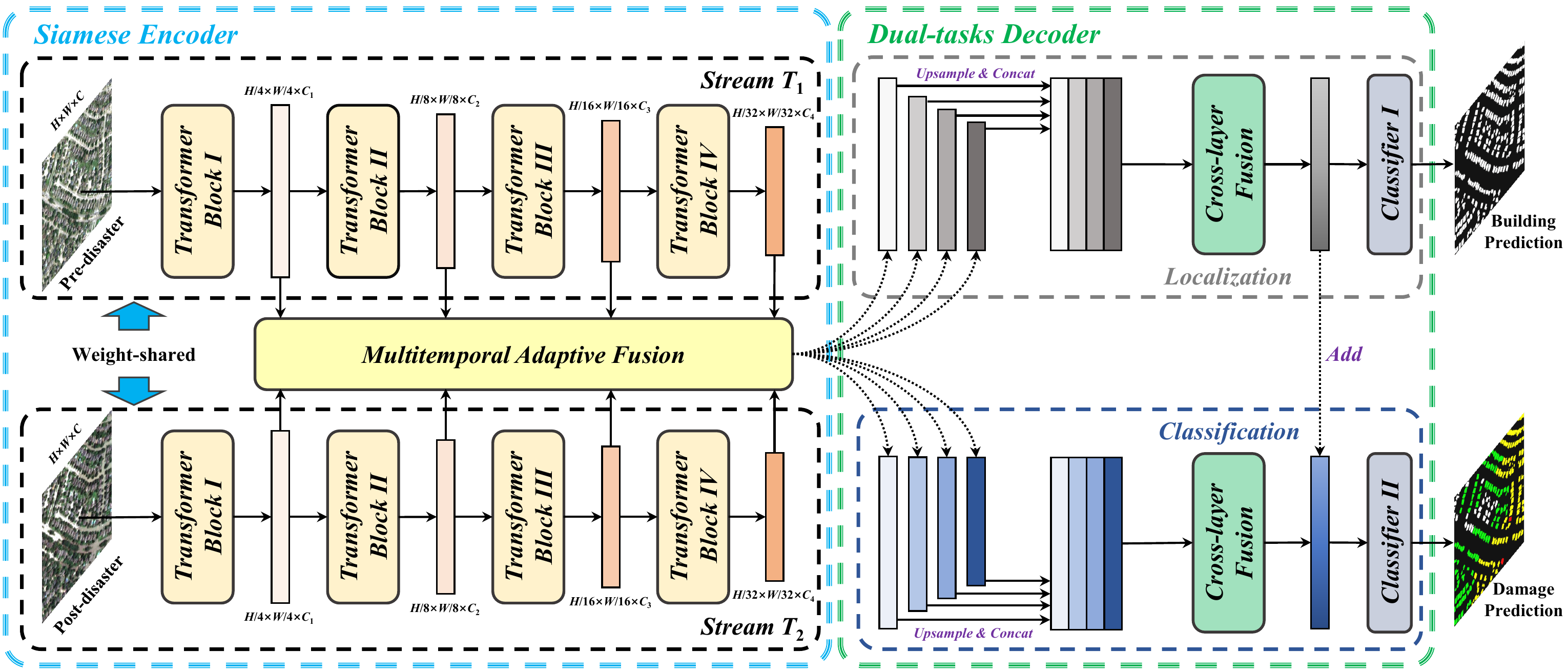}
  \caption{Overview of the proposed DamFormer architecture. }
  \label{fig:DamFormer}
\end{figure*}

The building damage assessment task is two-fold:  building localization and building per-pixel damage segmentation, namely damage classification. In this work, we present an end-to-end Transformer-based architecture DamFormer for both tasks. The overall proposed network architecture DamFormer is presented in Fig. \ref{fig:DamFormer}, which is constructed by two parts: a siamese Transformer encoder and a lightweight dual-tasks decoder.  

\subsection{Siamese Transformer Encoder}\label{sec:2.1}
\par Transformer backbones have shown their potential in feature extraction for semantic segmentation. Nevertheless, many of the current models are based on the ViT backbone, whose numerous network parameters inevitably introduce high computation overhead, which is paramount in the case of processing large-scale multitemporal remote sensing data. Therefore, in our work, we construct our siamese Transformer encoder based on the cutting-edge Transformer architecture SegFormer \cite{xie2021segformer}.

\par SegFormer consists of two main modules: a hierarchical transformer encoder named Mix Transformer (MiT) that outputs multi-level features and a lightweight All-MLP decoder to aggregate them and predict the semantic segmentation mask. As in ViT, an input image is first divided into patches, however MiT selects a smaller patches of size  4$\times$4 to favor the dense prediction task.  To reduce the computational complexity of the standard self-attention mechanism, MiT uses a sequence reduction process to reduce the length of the semantic sequence. Also, considering the positional encoding in Transformer demands the same resolution for input data in training and testing stages, a 3$\times$3 convolutional layer is introduced to replace the positional encoding. Benefiting from it, SegFormer architecture can accept multitemporal images of any size in training and testing stages. 

\par Specifically, based on the Mix Transformer encoder, our encoder has two streams to extract low-resolution fine-grained and high-resolution coarse features from multitemporal very high-resolution imagery where each stream has four Transformer blocks as shown in Fig. \ref{fig:DamFormer}. In this work, since pre-and post-disaster images are homogeneous with similar visual patterns, a weight-sharing mechanism is applied for the two streams, making the extracted features comparable (in the same feature space) and reducing network parameters. In addition, the overlap patch merging mechanism \cite{xie2021segformer} is introduced at each block to shrink the features obtained from the previous block to half the size. 

\par Finally, we propose a multitemporal adaptive fusion module, consisting of concatenation, convolutional layer, and channel attention mechanism \cite{woo2018cbam}. The concatenation operation and convolutional layer merge the feature maps of the same size from the two streams, which can facilitate the information interaction between pre-disaster and post-disaster feature maps. The channel attention mechanism can weight the importance of each channel and yield task-specific features that are well-suited for downstream building localization and damage classification tasks, respectively.

\subsection{Lightweight Dual-Tasks Decoder}\label{sec:2.2}
\par Keeping a large receptive field to include enough contextual information is key to segmenting remote sensing data. Therefore, sophisticated and computational demanding decoder designs such as multi-scale dilated convolution and self-attention head are often used by the CNN-based architecture. By contrast, our architecture can extract non-local features in each layer benefiting from the Transformer-based encoder. Hence, a simple lightweight decoder is therefore sufficient for downstream tasks.

\par Since the building damage assessment includes both building localization and damage classification tasks, our decoder includes two sub-networks with the same structure containing a cross-level fusion module and a classifier. In each decoder, the hierarchical task-specific features  are first upsampled to the same size. Then, the feature maps are concatenated and a cross-layer fusion model based on 
a 1$\times$1 convolutional layer is applied to aggregate information from different layers. The yielded feature map containing multi-level information is used for final prediction by an attached 1$\times$1 convolutional layer. In addition, the multi-level feature map in the localization sub-network is added back to the classification sub-network to contribute to the damage assessment performance. 
\begin{figure}[t]
  \centering
  \subfloat[]{
    \includegraphics[width=0.77in]{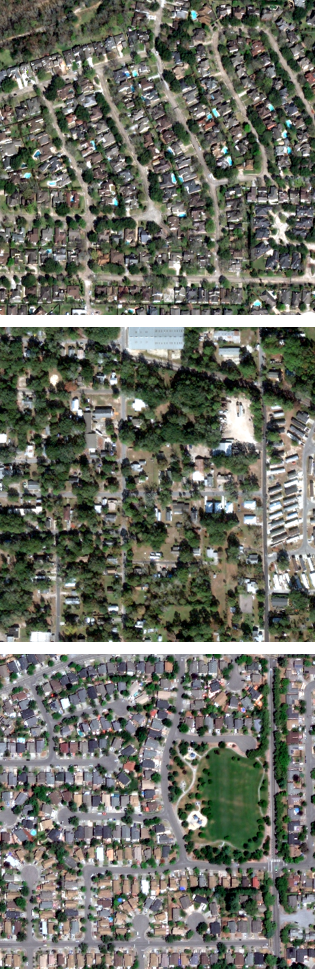}
  \label{PreImg}}
  \hfil
  \subfloat[]{
    \includegraphics[width=0.77in]{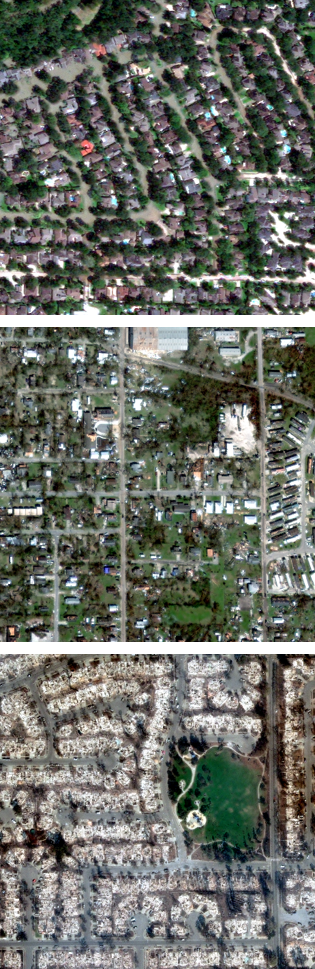}
  \label{PostImg}}
  \hfil
   \subfloat[]{
    \includegraphics[width=0.77in]{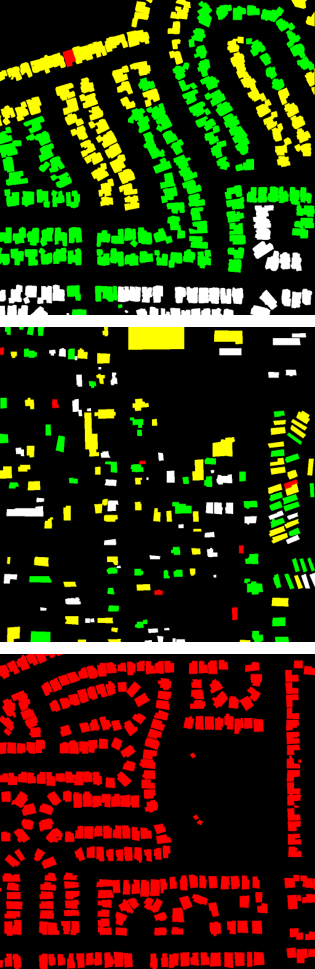}
  \label{Ref}}
  \hfil
  \subfloat[]{
    \includegraphics[width=0.77in]{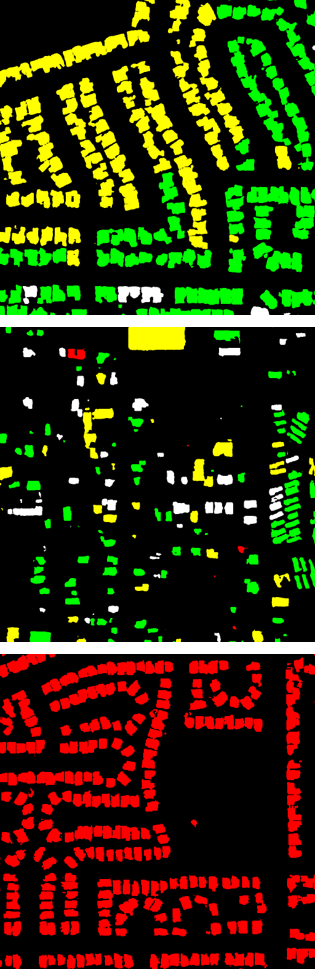}
  \label{Pre}}
  \caption{Visualization of the damage assessment results obtained by DamFormer from the xBD dataset. (a) Pre-disaster image. (b) Post-disaster image. (c) Reference map. (d) Prediction map. In (c) and (d), background pixels are shown in black, undamaged pixel in white, minor damaged pixels in green, major damaged pixels in yellow, and destroyed pixels in red.}
  \label{fig:2}
\end{figure}

\subsection{Loss Function}\label{sec:2.3}
\par To train the proposed architecture, we utilize a compound loss function to jointly optimize building localization and damage classification tasks. For building detection, we optimize predictions according to the building reference maps with binary cross-entropy loss. To address the problem of skewed class where pixels belonging to building classes (foreground pixels) are far less than background pixels, dice loss is introduced to balance the foreground and background pixels.
For damage assessment, we optimize predictions according to the damage reference maps with cross-entropy loss. Compared to building detection, the sample imbalance problem is more challenging in damage classification, which is not limited to the foreground and background pixels but also lies in the distribution of pixels with different damage levels. Considering this, we apply Lovasz softmax loss function \cite{berman2018lovasz} to support the optimization step, since it encounters sample imbalance problems by directly optimizing the IoU metric. 
Finally, our overall loss function can be formed as follows:

\begin{equation}
  L_{overall}=L_{loc}+\alpha L_{dam}
  \label{eq:1}    
\end{equation}
where $\alpha$ is a trade-off parameter that controls the importance of damage assessment, we set $\alpha$ to 1 in this paper. 

\begin{table*}[ht]\small
  \renewcommand{\arraystretch}{1.2}
  \caption{BENCHMARK COMPARISON OF DAMAGE ASSESSMENT RESULTS PRODUCED ON THE XBD DATASET}
  \label{table_1}
  \centering
  \begin{tabular}{c c c c c c c c}
    \hline
    \multirow{2}{*}{\textbf{Method}} & \multirow{2}{*}{\textbf{$F_{1}^{oa}$}}
    & \multirow{2}{*}{\textbf{$F_{1}^{loc}$}}  & \multirow{2}{*}{\textbf{$F_{1}^{dam}$}} &\multicolumn{4}{c}{\textbf{Damage $F_{1}$ per class}}   \\
    \cline{5-8} 
    &  & & & \textbf{No} &  \textbf{Minor} &  \textbf{Major} & \textbf{Destroyed}  \\
    \hline\hline
    xView2 Baseline &	26.54&	80.47&	3.42&	66.31&	14.35&	0.94&	46.57 \\ 	
    Siamese-UNet &	71.68&	85.92&	65.58&	86.74&	50.02&	64.43&	71.68\\ 
    MaskRCNN &	74.10&	83.60&	70.02&	\textbf{90.60}&	49.30&	72.20&	\textbf{83.70}\\ 
    ChangeOS&	75.50&	85.69&	71.14&	89.11&	53.11&	72.44&	80.79\\ 
    \textbf{DamFormer}&	\textbf{77.02}&	\textbf{86.86}&	\textbf{72.81}&	89.86&	\textbf{56.78}&	\textbf{72.56}&	80.51\\ 
    \hline
  \end{tabular}
\end{table*}

\section{Experiments}
\label{sec:experiments}
\subsection{Dataset, Metrics, Benchamrk Methods}\label{sec:general}

\par To evaluate the effectiveness of our DamFormer architecture, we conduct experiments on the xBD dataset \cite{gupta2019xbd}, which is currently the largest publicly available damage assessment dataset, containing 11'034 multitemporal very-high-resolution image-pairs with a size of 1024$\times$1024 and six disaster types: earthquake/tsunami, flood, volcanic eruption, wildfire, and wind. The damage annotations in the xBD dataset are divided into four levels: non-damage, minor damage, major damage, and destroyed. 
We implemented our network using Pytorch. In the two siamese streams, the specific number of MixFomer units in the four blocks is 3, 4, 6 and 3, keeping the same number of the residual units in ResNet-50 architecture. To train the overall network, we apply AdamW optimizer with an initial learning rate of 6$e^{-5}$ and a weight decay of 5$e^{-3}$.
\par Following the widely used metrics suggested in the xView2 Computer Vision for Building Damage Assessment
Challenge\footnote{https://www.xview2.org/}, an evaluation metric based on $F_{1}$ score was utilized, including building localization score $F_{1}^{loc}$, the harmonic mean of subclass-wise damage classification scores $F_{1}^{dam}$ and overall score $F_{1}^{oa}$, which can be formed as $F_{1}^{oa}=0.3 F_{1}^{loc}+0.7F_{1}^{dam}$. 
\par To evaluate the proposed framework we adopted four CNN-based architectures as a comparison methods, i.e., the  xView2 baseline method based on UNet + ResNet\footnote{https://github.com/DIUx-xView/xView2\_baseline}, the xView2 1st place solution method\footnote{https://github.com/DIUx-xView/xView2\_first\_place} Siamese-UNet, MaskRCNN-DA \cite{weber2020building}, and ChangeOS \cite{ZHENG2021112636}. Because the 1st place solution is based on a multi-model ensemble and ChangeOS applied object-based post-processing to improve the final performance, we used the ResNet-34 based models and the pixel-based ChangeOS respectively for a fair comparison.

\subsection{Experimental Results}\label{sec:experi_result}
\par Fig. \ref{fig:2} illustrates some visualization results obtained by DamFormer on the xBD dataset, which shows the proposed method can yield accurate and intact segmentation maps reflecting different levels of damages. Furthermore, in Table \ref{table_1}, we report the benchmark comparison on the xView2 \emph{holdout} split. As is evident, DamFormer outperformed on both building localization and damage classification tasks, which indicates the effectiveness of Transformer architecture in multitemporal remote sensing image processing tasks of building damage assessment. 

\section{Conclusion}
\label{sec:conclusion}

\par In this paper, we preliminary evaluate the potential of Transformer in multitemporal remote sensing data processing tasks. Specifically, we propose a dual-tasks siamese Transformer framework called DamFormer for building damage assessment tasks. DamFormer is made up of a siamese Transformer encoder and a lightweight dual-tasks decoder. Different from the limited receptive field of CNN-based architecture, DamFormer can extract non-local and representative features for building localization and damage assessment tasks. The experimental results on the xBD dataset demonstrate the superiority of our architecture for building damage assessment in comparison with CNN-based architectures. Our future work includes but is not limited to applying DamFormer architecture and its variants to man-made disasters response and further exploring Transfomer architecture in other remote sensing-related tasks, such as change detection, land-cover, and land-use classification.


\small
\bibliographystyle{IEEEbib}
\bibliography{Template}

\begin{thebibliography}{10}

\bibitem{rs12162532}
E.~Nemni, J.~Bullock, S.~Belabbes, and L.~Bromley,
\newblock ``Fully convolutional neural network for rapid flood segmentation in
  synthetic aperture radar imagery,''
\newblock {\em Remote Sens.}, vol. 12, no. 16, 2020.

\bibitem{Chen2020TGRS}
H.~Chen, C.~Wu, B.~Du, L.~Zhang, and L.~Wang,
\newblock ``Change detection in multisource vhr images via deep siamese
  convolutional multiple-layers recurrent neural network,''
\newblock {\em IEEE Trans. Geosci. Remote Sens.}, vol. 58, no. 4, pp.
  2848--2864, 2020.

\bibitem{ZHENG2021112636}
Z.~Zheng, Y.~Zhong, J.~Wang, A.~Ma, and L.~Zhang,
\newblock ``Building damage assessment for rapid disaster response with a deep
  object-based semantic change detection framework: From natural disasters to
  man-made disasters,''
\newblock {\em Remote Sens. Environ.}, vol. 265, pp. 112636, 2021.

\bibitem{chen2017deeplab}
L.~Chen, G.~Papandreou, I.~Kokkinos, K.~Murphy, and A.~L. Yuille,
\newblock ``Deeplab: Semantic image segmentation with deep convolutional nets,
  atrous convolution, and fully connected crfs,''
\newblock {\em IEEE Trans. Pattern Anal. Mach. Intell.}, vol. 40, no. 4, pp.
  834--848, 2017.

\bibitem{zhang2019self}
H.~Zhang, I.~Goodfellow, D.~Metaxas, and A.~Odena,
\newblock ``Self-attention generative adversarial networks,''
\newblock in {\em ICML}, 2019, pp. 7354--7363.

\bibitem{vaswani2017attention}
A.~Vaswani, N.~Shazeer, N.~Parmar, J.~Uszkoreit, L.~Jones, A.~N. Gomez,
  {\L}.~Kaiser, and I.~Polosukhin,
\newblock ``Attention is all you need,''
\newblock in {\em NIPS}, 2017, pp. 5998--6008.

\bibitem{dosovitskiy2021an}
A.~Dosovitskiy, L.~Beyer, A.~Kolesnikov, D.~Weissenborn, X.~Zhai,
  T.~Unterthiner, M.~Dehghani, M.~Minderer, G.~Heigold, S.~Gelly, J.~Uszkoreit,
  and N.~Houlsby,
\newblock ``An image is worth 16x16 words: Transformers for image recognition
  at scale,''
\newblock in {\em ICLR}, 2021.

\bibitem{xie2021segformer}
E.~Xie, W.~Wang, Z.~Yu, A.~Anandkumar, J.~M. Alvarez, and P.~Luo,
\newblock ``Segformer: Simple and efficient design for semantic segmentation
  with transformers,''
\newblock {\em arXiv preprint arXiv:2105.15203}, 2021.

\bibitem{Liu_2021_ICCV}
Z.~Liu, Y.~Lin, Y.~Cao, H.~Hu, Y.~Wei, Z.~Zhang, S.~Lin, and B.~Guo,
\newblock ``Swin transformer: Hierarchical vision transformer using shifted
  windows,''
\newblock in {\em ICCV}, October 2021, pp. 10012--10022.

\bibitem{woo2018cbam}
S.~Woo, J.~Park, J.~Lee, and I.~Kweon,
\newblock ``Cbam: Convolutional block attention module,''
\newblock in {\em ECCV}, 2018, pp. 3--19.

\bibitem{berman2018lovasz}
M.~Berman, A.~R. Triki, and M.~B. Blaschko,
\newblock ``The lov{\'a}sz-softmax loss: A tractable surrogate for the
  optimization of the intersection-over-union measure in neural networks,''
\newblock in {\em CVPR}, 2018, pp. 4413--4421.

\bibitem{gupta2019xbd}
R.~Gupta, R.~Hosfelt, S.~Sajeev, N.~Patel, B.~Goodman, J.~Doshi, E.~Heim,
  H.~Choset, and M.~Gaston,
\newblock ``xbd: A dataset for assessing building damage from satellite
  imagery,''
\newblock {\em arXiv preprint arXiv:1911.09296}, 2019.

\bibitem{weber2020building}
E.~Weber and H.~Kan{\'e},
\newblock ``Building disaster damage assessment in satellite imagery with
  multi-temporal fusion,''
\newblock {\em arXiv preprint arXiv:2004.05525}, 2020.

\end{thebibliography}

\end{document}